\pdfoutput=1

\documentclass[11pt]{article}

\usepackage[]{EMNLP2022}

\usepackage{times}
\usepackage{latexsym}

\usepackage[T1]{fontenc}

\usepackage[utf8]{inputenc}

\usepackage{microtype}

\usepackage{inconsolata}
\usepackage{graphicx}
\usepackage{xcolor}         

\usepackage{caption}
\usepackage{subcaption}
\definecolor{mygreen}{RGB}{50, 128, 50}

\definecolor{myorange}{RGB}{255, 165, 0}

\usepackage[normalem]{ulem}
\useunder{\uline}{\ul}{}
\usepackage{lscape}
\usepackage{longtable}
\usepackage{booktabs}
\usepackage{float}
\restylefloat{table}

\newcommand{\compoundvar}[1]{{\color{red}#1}}
\newcommand{\ournl}{\textbackslash n}
\newcommand{\nlnl}{\ournl\ournl}
\newcommand{\promptenv}[1]{{\raggedright\texttt{#1}}}
\newcommand{\word}[1]{\emph{#1}}

\usepackage{colortbl}
\newcolumntype{P}[1]{>{\raggedright\arraybackslash}p{#1}}
\title{Systematicity in GPT-3's Interpretation of Novel English Noun Compounds\thanks{{}~Experimental materials available at \url{https://github.com/siyan-sylvia-li/systematicity_gpt3/}}}

\author{Siyan Li \\
  Stanford University \\
  \texttt{siyanli@stanford.edu} \\\And
  Riley Carlson \\
  Stanford University \\
  \texttt{rileydc@stanford.edu}  \\\And
  Christopher Potts \\
  Stanford University \\
  \texttt{cgpotts@stanford.edu} \\}

\begin{document}
\maketitle
\begin{abstract}
\citet{levin2019systematicity} show experimentally that the interpretations of novel English noun compounds (e.g., \emph{stew skillet}), while not fully compositional, are highly predictable based on whether the modifier and head refer to artifacts or natural kinds. Is the large language model GPT-3 governed by the same interpretive principles? To address this question, we first compare \citeauthor{levin2019systematicity}'s experimental data with GPT-3 generations, finding a high degree of similarity. However, this evidence is consistent with GPT-3 reasoning only about specific lexical items rather than the more abstract conceptual categories of \citeauthor{levin2019systematicity}'s theory. To probe more deeply, we construct prompts that require the relevant kind of conceptual reasoning. Here, we fail to find convincing evidence that GPT-3 is reasoning about more than just individual lexical items. These results highlight the importance of controlling for low-level distributional regularities when assessing whether a large language model latently encodes a deeper theory.
\end{abstract}

\section{Introduction}

English noun compounds (e.g., \emph{birthday cake}, \emph{toy store}) have long been central to debates about the limits of compositionality in natural language semantics. For \citet{Partee95}, they are essentially memorized lexical items whose meanings cannot be predicted from their parts. In contrast, \citet{downing1977creation}, \citet{wisniewski1998relations}, and \citet{levi1978syntax} identify a range of systematic constraints on compound interpretation, suggesting underlying cognitive and linguistic properties that are more complex and productive than memorization would suggest. More recently, \citet{levin2019systematicity} show that, for large classes of noun compounds, the meanings are highly predictable. Their evidence comes from production and comprehension experiments with novel noun compounds (e.g., \emph{stew skillet}, \emph{swamp squash}), for which participant interpretations are highly consistent. These results suggest that, while noun compound meanings are not fully compositional, they are \emph{systematic} in the sense of \citealt{Fodor:Pylyshyn:1988}.

In this paper, we ask whether the large language model (LLM) GPT-3 \citep{brown2020language} is similarly systematic in its handling of novel compounds. Our comparisons focus on the free-response comprehension experiments of \citet{levin2019systematicity}.  \citeauthor{levin2019systematicity}'s guiding hypothesis is that the observed systematicity traces to two over-arching conceptual hypotheses. The \emph{Event-Related Hypothesis} says that, for compounds referring to artifacts (e.g., \emph{stew skillet}), the modifier will tend to convey information about the creation or use of the compound's referent. The \emph{Essence-Related Hypothesis} says that, for compounds referring to natural kinds (e.g., \emph{swamp squash}), the modifier will tend to identify essential properties of the compound's referent. In other words, deep conceptual properties of the component words are constraining how people interpret novel compounds. Is GPT-3 governed by the same interpretive principles? We report on three experiments seeking to address this question.

In Experiment~1, we find that, when prompted with novel compounds, GPT-3's behavior is strikingly aligned with the interpretations human participants gave in \citeauthor{levin2019systematicity}'s experiments. In Experiment~2, we employ the same protocol but with new noun compounds involving a wider range of modifier--head relationships, and we see essentially the same behavior from GPT-3. These results are consistent with the claim that GPT-3 is governed by the Event- and Essence-Related Hypotheses, but it is far from conclusive, since the model may simply be relying on regularities in the interpretations of individual words, rather than reasoning about artifacts or natural kinds as abstract conceptual categories. In Experiment~3, we seek to decouple these two explanations by constructing prompts in which random strings are defined as natural kinds or artifacts and then used in novel compounds. In this setting, GPT-3 is much less successful, suggesting that Experiments~1 and 2 may have confounded statistical distributions of tokens with deeper conceptual understanding. Overall, this highlights the importance of controlling for low-level distributional regularities when assessing whether an LLM latently encodes a deeper conceptual or linguistic theory.

\begin{table*}[!tbh]
    \centering
    \begin{tabular}{ll}
    \toprule
    \textbf{Meta-relation} & \textbf{Subtypes}\\
    \midrule    
    Event & made of, method, purpose, time, used by, object-nom \\
    Essence & borrowed, color, dimension, distinctive part, taste/smell, location, social/political \\   
    Other & named after, value, whole-part, other\\
    \bottomrule
    \end{tabular}
    \caption{Meta-relations and their corresponding subtypes present in this study. Complete definitions are given in \citealt{levin2019systematicity}, Appendix~A.}
    \label{tab:meta}
\end{table*}

\section{Background}

\subsection{English Noun Compounds}

We follow \citet{levin2019systematicity} in focusing on English \emph{endocentric} noun compounds like \word{soup spoon} and \word{string bean}. For our purposes, the crucial feature of endocentric compounds is that they entail their head noun: a soup spoon is a spoon, a string bean is a bean, and so forth. By contrast, \emph{exocentric} compounds like \word{ladyfinger} and \word{paperback} do not entail either of their component parts (e.g., a ladyfinger is neither a lady nor a finger, but rather a sweet treat). Exocentric compounds may simply be memorized lexical items.

\citealt{downing1977creation} is a groundbreaking study that uses the production and comprehension of novel compounds to explore the systematicity of noun compounds. \citeauthor{downing1977creation} proposes that compounds are devices for communicating about objects by identifying their salient features. Inspired by \citet{downing1977creation}'s suggestions, \citet{levi1978syntax} studies recovery of deleted information in dialogue. Specifically, when compounds are created to describe objects, the relationship between the compound head and modifier is often omitted. \citeauthor{levi1978syntax} formalizes nine semantic categories for relationships between compound heads and modifiers. \citet{wisniewski1998relations} examine noun--noun compounds describing office supplies versus wildlife, and discover systematic differences in the relationships between compound heads and modifiers for these two types of entities.

\subsection{LLMs and Linguistic Creativity}

LLMs have been assessed in a range of tasks involving constrained creativity with language. 

\citet{malkin-etal-2021-gpt} show that GPT-3 can define novel nonsense words in ways that seem plausible to human evaluators. Similarly,  \citet{pinter-etal-2020-will} study blends like \emph{thruple} (`three-person couple'), alongside more transparent cases (e.g., \emph{quiz-maker}) and more opaque ones (\emph{deathbox}, `dangerous car').

\citet{chakrabarty2021s} find that enhancing GPT-2 with contextual or literal knowledge outperforms few-shot GPT-3 when continuing figurative narratives containing idioms and similes. 

\citet{yu-etal-2020-homophonic} generate homophonic puns using a constraint selection process that rewrites sentences into puns in a semantically naturalistic manner, while \citet{mittal2022ambipun} first generate context words related to different pun word senses through GPT-3, then combine context words with separate word senses to create homographic puns.

Idioms provide an interesting comparison with noun compounds, since we do not expect a high degree of predictability for them (though idioms do show some aspects of compositonality; \citealt{Nunberg:Sag:Wasow:1994}). \citet{socolof-etal-2022-characterizing} use BERT and XLNet to develop contextual metrics for idiom classification, and \citet{tan-jiang-2021-bert} probe BERT's and ERNIE2's capacities to distinguish literal and idiomatic uses of a potentially idiomatic expression, and to identify the proper paraphrases of idiomatic expressions in a given context.

Novel noun compounds have the potential to offer important new insights in this area. First, unlike novel blends and idioms, novel compounds are easy to create, and \citeauthor{levin2019systematicity}'s findings indicate that we can expect high predictability for them. Second, \citeauthor{levin2019systematicity}'s hypotheses engage directly with deeper cognitive notions, rather than being purely about linguistic forms.

\section{Experiment 1: Free-Response with \citeauthor{levin2019systematicity}'s Novel Compounds}

In \citeauthor{levin2019systematicity}'s novel compound comprehension study,\footnote{Data available at \url{https://osf.io/t43kd/}} participants gave free-form textual responses to prompts like ``Imagine that you encounter the compound \texttt{X}. What would you think this refers to?'' Expert labelers then annotated each response for its metarelation (relationship between the compound head and its modifier) and metarelation subtype, using definitions summarized in Table~\ref{tab:meta}. The final experimental dataset consists of 798 participant-created explanations for 38 different novel compounds. 141 explanations were excluded in accordance with the coders' manual.

\subsection{Methods}

Our experiment essentially treats GPT-3 (specifically, the Instruct-GPT Davinci model) as a new participant in this experiment. For each compound, we create three separate prompts:
\begin{description}\setlength{\itemsep}{0pt}
\item[Natural:] 
\promptenv{Imagine that you encounter the compound \compoundvar{X}. What would you think this refers to?}  

\item[Structured:]
\promptenv{Compound:\,\compoundvar{X}.\ournl\ournl Explanation:}

\item[Few-shot]
\promptenv{%
Compound:\,\compoundvar{X$_{1}$}.\nlnl
Explanation:\,\compoundvar{E$_{1}$}\nlnl\\
Compound:\,\compoundvar{X$_{2}$}.\nlnl
Explanation:\,\compoundvar{E$_{2}$}\nlnl\\
Compound:\,\compoundvar{X$_{3}$}.\nlnl
Explanation:\,\compoundvar{E$_{3}$}\nlnl\\
Compound:\,\compoundvar{X}.\nlnl
Explanation:}   
\end{description}

In the few-shot prompt, the three examples are randomly chosen from the novel compounds in \citealt{levin2019systematicity}. Given a randomly selected compound \compoundvar{X$_{i}$}, we then sample, from \citeauthor{levin2019systematicity}'s data, a human-generated explanation \compoundvar{E$_{i}$} from the majority metarelation for \compoundvar{X$_{i}$}.

We obtain top-1 samples from the model with temperatures 0.2, 0.4, 0.7 and 0.9, to simulate different response behaviors, from almost deterministic to highly variable. Combining the temperatures and prompts results in 12 conditions for each of the 38 compounds, for a total of 456 generations. Each generation was annotated by each of us (the authors) using the coding framework of \citeauthor{levin2019systematicity}. We saw only the compound and the generated text; the prompt and temperature value were hidden. The Fleiss' kappa for these annotations is 0.844 at the level of meta-relations and 0.767 at the level of relation sub-types.

\begin{figure}[tp]
    \centering
    \includegraphics[width=\linewidth]{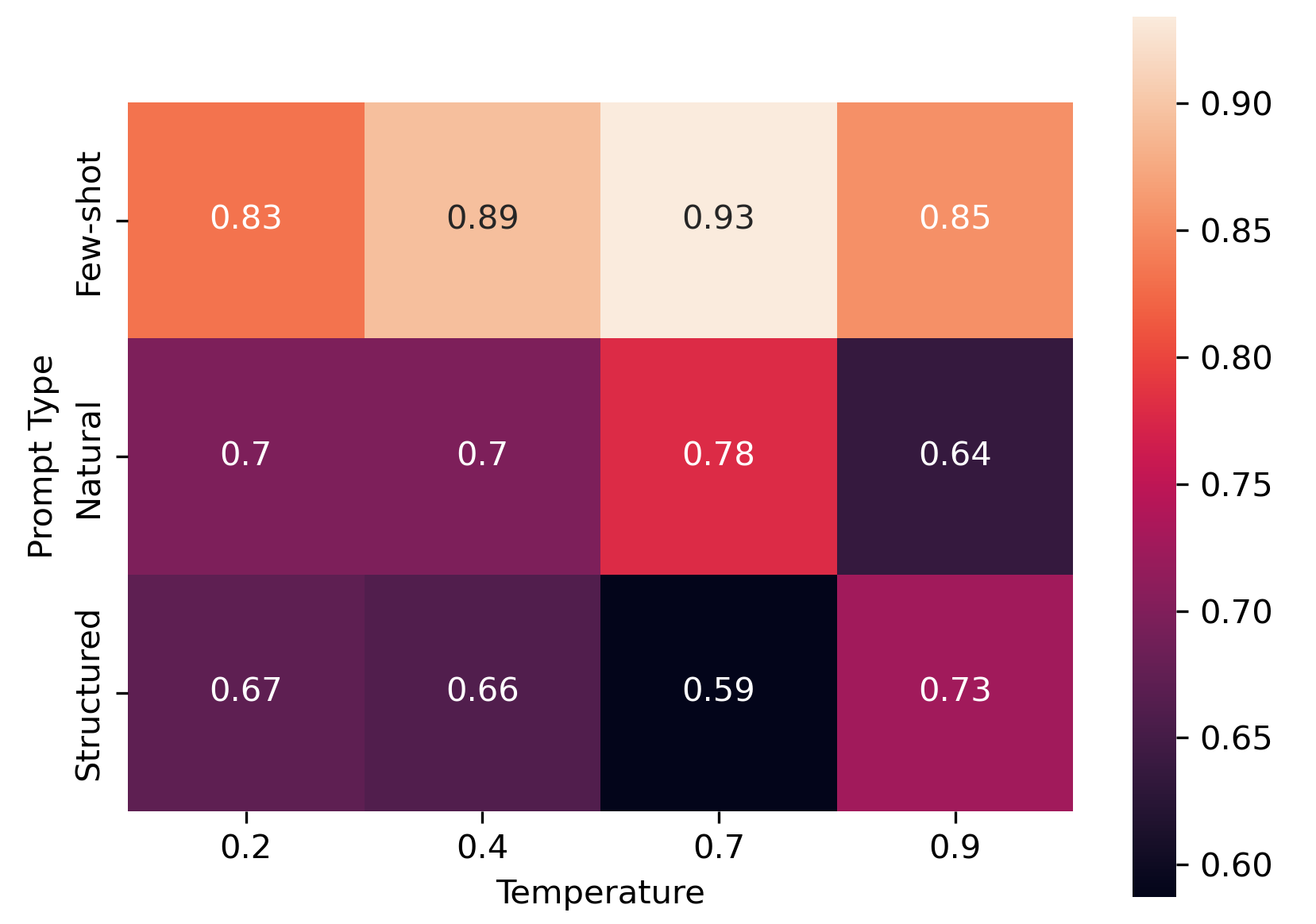}
    \caption{Experiment~1 correlations between model generations' metarelation distributions and \citet{levin2019systematicity} metarelation distributions. Comparatively, temperature 0.7 with few-shot prompting appears optimal.}
    \label{fig:heatmap}    
\end{figure}

\subsection{Results}

Figure~\ref{fig:heatmap} provides the correlations with \citet{levin2019systematicity}'s human response data for our three prompt types and four temperature settings. The correlations are between metarelation distributions from the human comprehension study and coded GPT-3 generations. We provide more detail on generating the figure in Appendix~\ref{app:experiment1}. On a high level, \citet{levin2019systematicity} computed the percentages of coded metarelations for different noun compound types based on the semantic property of the heads and the modifiers. Specifically, the novel compounds are divided into four categories based on whether a compound component is a natural kind or is artificial. We compute the percentages of metarelations exhibited by GPT-3 generations, and average the Pearson's correlations of each compound category between our percentages and those of \citet{levin2019systematicity}. All of the correlations are very high, with the few-shot prompt and 0.7 temperature parameter giving the highest correlation, at 0.93. Overall, these results indicate that GPT-3 is returning definitions whose interpretations align with those given by humans.

Another important dimension is the rate at which definitions need to be excluded because they do not conform to the constraints of endocentric compounds or are otherwise unintelligible. Figure~\ref{tab:my_label} provides the exclusion rates for all our model variants and \citet{levin2019systematicity}'s human response data. GPT-3's exclusion rates are actually much lower than the human rate, suggesting that, in some sense, it may be \emph{more} systematic than humans when confronted with novel compounds.

\section{Experiment~2: Free-Response with New Novel Compounds}

We are unable to determine whether \citet{levin2019systematicity}'s experimental data might have been part of GPT-3's training or fine-tuning data. In addition, their set of compounds covers a relatively small part of the full space defined by the coding manual, which might make the problem artificially easy for the model. To address both of these concerns, we created a new set of compounds that keyed into different parts of Table~\ref{tab:meta}, and we repeated our experiment with GPT-3 using this new set.

The additional compounds do not have any lexical overlap with those of  \citealt{levin2019systematicity}. Furthermore, we observed in the previous experiment that the generations for \citet{levin2019systematicity}'s novel compounds are categorized primarily as the ``purpose'' metarelation subtype. Therefore, we also include compounds that, according to our intuitions, should have more varied metarelation subtypes. We use the same temperatures and prompts combinations.

Table~\ref{tab:exp2-exclude} provides the exclusion percentages for this experiment. The rates are overall slightly higher than for Experiment~1, but they are still strikingly low, especially for the few-shot prompt. The Fleiss kappas for the annotations are also very similar to Experiment~1: 0.786 for meta-relations and 0.714 for relation sub-types.  Appendix~\ref{app:additional} provides additional details on this experiment.

\begin{figure}[tp]    
    \centering
    \begin{tabular}[b]{r rrrr}
    \toprule
        & \multicolumn{4}{c}{\emph{Temperature}} \\
        & \textbf{0.2} & \textbf{0.4} & \textbf{0.7} & \textbf{0.9} \\
         \midrule
       \textbf{Natural}  & 13.15 & 10.52 & 21.05 & 28.94\\
       \textbf{Structured} & 28.94 & 36.84 & 28.94 & 34.21 \\
       \textbf{Few-shot} & 0.0 & 0.0 & 5.26 & 2.63 \\
       \midrule
       \textbf{Human} & \multicolumn{4}{c}{17.67}\\
       \bottomrule
    \end{tabular}
    \caption{Experiment~1 exclusion percentages per prompt/temperature combination. We also report the percentage of human-generated responses that were excluded in \citet{levin2019systematicity}.}
    \label{tab:my_label}
\end{figure}

\begin{figure}[tp]
    \centering
    \begin{tabular}{*{5}{r} }
    \toprule
         & \multicolumn{4}{c}{\emph{Temperature}} \\
         & \textbf{0.2} & \textbf{0.4} & \textbf{0.7} & \textbf{0.9} \\
         \midrule
       \textbf{Natural}  & 5.56 & 11.43 & 5.56 & 5.88\\
       \textbf{Structured} & 14.29 & 17.14 & 19.44  & 44.12 \\
       \textbf{Few-shot} & 5.71 & 5.71  & 11.43 & 8.57 \\
       \bottomrule
    \end{tabular}
    \caption{Experiment~2 exclusion percentages per prompt/temperature combination.}
    \label{tab:exp2-exclude}
\end{figure}  

\section{Experiment 3: In-Context Conceptual Reasoning}

The results of Experiments~1 and 2 are consistent with GPT-3's behavior being governed by the Event- and Essence-Related Hypotheses at some level, but many other explanations are available. In particular, it could very well be that GPT-3 is reasoning about individual lexical items without truly being able to identify them as artifacts or natural kinds or relate such classifications to its overall generations. To address this confound, we conducted a third experiment.

\subsection{Methods}

For this experiment, the target compound is created from the nonsense strings \texttt{gmtomflxri} and \texttt{putrlv} using a prompt that requires implicit inferences about the artifact vs.~natural kind status of these terms based on their in-context definitions:
\begin{quote}
\promptenv{A \compoundvar{modifier} is \compoundvar{definition$_{1}$}.\\ 
A \compoundvar{head} is \compoundvar{definition$_{2}$}.\\
A \compoundvar{compound} is \compoundvar{definition$_{3}$}.\nlnl\\    
A gmtomflxri is \compoundvar{definition$_{4}$}.\\ 
A putrlv is \compoundvar{definition$_{5}$}.\\
A gmtomflxri putrlv is}
\end{quote}
We used pure nonsense strings to avoid inadvertently using nonce words that the model might have seen in other contexts. We use four noun compounds for each combination of artifact and natural kind for head and modifier categories. The basis compounds are \emph{kitchen knife}, \emph{tree frog}, \emph{strawberry cookie}, and \emph{coffee bean}. This results in a total of 16 new compounds per basis compound.

For each basis compound, we generate new compounds from the components with specific relationships to the original components. We define these relationships along two axes: (1) Match (M) or Different (D) in noun category, and (2) Close (c) or Far (f) in semantic space. For example, \emph{bedroom} is an Mc instance of \emph{kitchen} because they are both artifacts and similar in concept, and \emph{fur} is a Df instance of \emph{kitchen} because \emph{fur} is a natural kind and is not similar in concept. To decide semantic proximity, we utilize both WordNet \cite{wn} for Match instances and GloVe embeddings \cite{pennington2014glove} for Different instances. (See Appendix~\ref{app:proximity} for details.)

For a concrete example of our prompting process, assume that we select the basis compound \emph{strawberry cookie} and the relationship Mc/Mf. Using the pre-defined list of basis compounds and their variations (Table~\ref{tab:full_alien} in Appendix~\ref{app:materials}), we select the Mc word of \emph{strawberry}, \emph{banana}, and the Mf word of \emph{cookie}, \emph{table}. Referencing the list of definitions corresponding to each word defined in the appendix, we have the following prompt:

\begin{quote}
\promptenv{A \compoundvar{strawberry} is \compoundvar{sweet fleshy red fruit}.\\ 
A \compoundvar{cookie} is \compoundvar{any of various small flat sweet cakes}.\\
A \compoundvar{strawberry cookie} is \compoundvar{a cookie made with strawberries}.\nlnl\\    
A gmtomflxri is \compoundvar{a tropical yellow fruit}.\\ 
A putrlv is \compoundvar{a piece of furniture}.\\
A gmtomflxri putrlv is}
\end{quote}

The model generations were annotated by us. The Fleiss' kappas are very similar to those of Experiments~1 and 2: 0.821 for the meta-relations and 0.803 for the subtypes.

\subsection{Results}

The first thing that stands out about these new results is that the exclusion rate is much higher. Table~\ref{tab:alien_exclude} shows the results for the temperature parameter of 0.2, which had the lowest exclusion rates overall. Approximately 30\% of the exclusions are inversions of heads and modifiers (e.g., defining \word{banana table} as a banana instead of as a table). This suggests inconsistent knowledge of this core structural distinction when presented with nonce words. It is unclear if such core structural knowledge is missing in Experiment~1 and 2, since the results may have been masked by lexical regularities. Other instances of exclusion include vague definitions, directly concatenating head and modifier definitions together without proper explanations, and copying the prompt (e.g., \word{cider nut} is often defined as \word{coffee bean} because \word{coffee bean} is a part of the prompt). 

\begin{table}[tp]
    \centering
    \begin{tabular}{rrrrr @{\hspace{18pt}} r}
    \toprule
        \small{\underline{\textit{Head}}} & \textbf{Mc} & \textbf{Mf} & \textbf{Dc} & \textbf{Df} & \textbf{Avg.}\\
        \midrule
        \small{\underline{\textit{Modifier}}} & & & & &\\
        \textbf{Mc} & 25.0 & 50.0 & 0.0 & 75.0 & 37.5\\
        \textbf{Mf} & 0.0 & 0.0 & 0.0 & 0.0 & 0.0\\
        \textbf{Dc} & 50.0 & 25.0 & 75.0 & 33.3 & 45.8 \\
        \textbf{Df} & 75.0 & 25.0 & 50.0 & 50.0 & 50.0
        \\[1ex]
        \textbf{Avg.} & 37.5&25 & 31.25 & 39.5 & \\
        \bottomrule
    \end{tabular}
    \caption{Exclusion percentage for every head-modifier combination in Experiment~2, with temperature 0.2.}
    \label{tab:alien_exclude}
\end{table}

For the explanations that were not excluded, we calculate the distributions of metarelation subtype coding per compound type, similar to \citet{levin2019systematicity}. The percentages of meta-relations per compound type are presented in Table~\ref{tab:alien_meta}. While there is limited correlation between compound type with the percentage of meta-relations coded as ``Events'', there does not appear to be as clear a correlation with the original study as in Experiment~1; the Pearson's correlation to the original distribution is only 0.105 (compared to 0.93 in Experiment~1). 

\begin{table}[tp]
    \centering
    \setlength{\tabcolsep}{5pt}
    \begin{tabular}{ *{6}{r} }
         \toprule
         \textbf{Type} & \textbf{Event} & \textbf{Percep.} & \textbf{Env.} & \textbf{Other} & \textbf{N}\\
         \midrule
         \textbf{art-art} &  35.1 & 8.1 & 56.8 & 0 & 37 \\
         \textbf{art-nk} &  46.9 & 0 & 53.1 & 0& 32\\
         \textbf{nk-art} &  73.9 & 17.4 & 6.5 &2.2 & 46\\
         \textbf{nk-nk} &  33.3 & 18.0 & 20.5 & 28.2& 39\\
         \bottomrule
    \end{tabular}
    \caption{Coded metarelation distribution per compound type for ``Alien'' compound experiments. ``Art'' means ``artifact'' and ``nk'' means ``natural kind''. ``Art-nk'' indicates the type of compounds with an artifact head and a natural kind modifier.}
    \label{tab:alien_meta}
\end{table}

\section{Discussion and Conclusion}

This paper explored GPT-3's handling of novel English endocentric noun compounds, building on psycholinguistic research by \citet{levin2019systematicity}. Experiments~1 and 2 suggested that GPT-3 might be governed by the same over-arching conceptual and linguistic principles that shape human interpretations of these forms. However, Experiment~3 probed that conclusion more deeply and failed to find evidence for this stronger claim: GPT-3 may instead be memorizing token distributions rather than reasoning about the underlying conceptual categories on which \citet{levin2019systematicity}'s theory depends.

\section{Limitations}

In Experiments~1 and 2, we assumed that the model already knows the distinction between compound heads and modifiers, yet in Experiment~3 we discovered that head--modifier reversal is a major factor in exclusions. Therefore, the assumption that GPT-3 inherently understands compound head versus modifier may be incorrect. This warrants a more thorough investigation, to examine GPT-3 generations of flipped versions of compounds (e.g., evaluate both \word{duck potato} and \word{potato duck}). Although the switching of the head and modifier was not random enough to suggest that the model has no understanding of noun compounds, this investigation could help determine whether the model inherently understands the relationship between the head and modifier.

Additionally, the statistical distribution of compound components in the training data of GPT-3 may be a confounding variable in our experiment results. However, we were unable to test this theory using statistical analyses due to the model's training data not being publicly available. If the statistical distribution of the compound components in the training data were a confounding variable, then it may be the case that the compounds used did not thoroughly probe the models understanding of noun compounds and that other compounds should be used. Experiment~2 reassures us somewhat, but it is not a substitute for a thorough audit of the training data.

The GPT-3 model annotations were done by the three authors of the paper. While it might seem better to train separate annotators, this is not a simple matter, as its requires deep linguistic expertise. \citet{levin2019systematicity} did much of their coding themselves, presumably for this reason. Importantly, when we did the coding, we were not able to see which model variant produced the generation being evaluated, which allowed for an unbiased comparison across model types even factoring in any annotator biases. In addition, we note that one of our experiments led to a positive result for GPT-3 and the other negative, consistent with us having no particular preferred outcome for the paper's findings. The high Fleiss kappa scores further support the claim that our work simply implemented the coding manual from \citealt{levin2019systematicity}. All our annotations are included in our public repository: \url{https://github.com/siyan-sylvia-li/systematicity_gpt3/}.

\bibliography{anthology,custom}
\bibliographystyle{acl_natbib}

\appendix

\section*{Appendix}

\begin{table*}[h!]
    \centering
    \setlength{\itemsep}{4pt}
    \begin{tabular}{
        P{0.15\textwidth}
        P{0.45\textwidth}
        P{0.3\textwidth}
    }
    \toprule
        \textbf{Compound} & \textbf{Excluded Explanation} & \textbf{Reason}\\
    \midrule
        panini leaf 
        & 
        A type of sandwich that is usually made with panini bread. & Defining only compound modifier\\
        \midrule
        stick broccoli 
        & 
        The word ``stick'' is a verb meaning to pierce or to fasten. In this sentence, it is used as a verb meaning to fasten. The word ``broccoli'' is a noun meaning a type of vegetable.
        & Defining each part separately\\
        \midrule
        stick broccoli 
        & 
        The term ``stick broccoli'' is not a real compound, so it is difficult to say what it would refer to. 
        & Not defining the compound \\
        \midrule
        stew chickpea 
        & 
        A stew made with chickpeas. & Defining wrong compound head \\
        \bottomrule
    \end{tabular}
    \caption{Example generations that are excluded.}
    \label{tab:exclude}
\end{table*}

\section{Experiment 1: Additional Details}\label{app:experiment1}

\subsection{Details for Generating the Correlation Heatmap}
The correlation heatmap in Figure~\ref{fig:heatmap} is created by calculating the average correlation between coded metarelations (not including subtypes) from \citet{levin2019systematicity} and GPT-3 generations. There are four noun compound types, and each of them has different percentages for event, location, and perceptual in both \citealt{levin2019systematicity} and our experiments. For a specific set of GPT-3 generations, we first compute the percentage of each metarelation for each of the four noun compound types. Then, for each noun compound type, we compute the Pearson's correlations between the metarelation percentage vector of the generations with the metarelation vector from \citet{levin2019systematicity}. Finally, we average the coefficients across the four compound types. This number is then reported as the correlation coefficient for that specific set, in this case a prompt/temperature combination. The distribution from \citet{levin2019systematicity} is in Table~\ref{tab:dists}.

\subsection{Examples of Experiment~1 Generations}

We present examples of both GPT-3 generations and human-created definitions from \citet{levin2019systematicity}:

\begin{center}
    \setlength{\tabcolsep}{2pt}
    \begin{tabular}{P{0.12\textwidth} P{0.33\textwidth}}
    \toprule
       \textbf{Compound} & \textbf{GPT-3 Generation} \\
       \midrule
        water spatula
       & 
       A water spatula is a kitchen utensil that is used to help transfer liquids from one container to another. ... \\
        \midrule        
        stew skillet
       & A skillet used to make stews. \\
       \midrule        
        duck screen
       & 
       A duck screen is a type of netting that is used to keep ducks and other waterfowl out of areas where they are not wanted. \\
     \bottomrule
    \end{tabular}
    \label{tab:example_gens}
\end{center}

\vspace{12pt}

\begin{center}
    \setlength{\tabcolsep}{2pt}
    \begin{tabular}{P{0.12\textwidth} P{0.33\textwidth}}
    \toprule
       \textbf{Compound} & \textbf{Human Definition} \\
       \midrule       
       water spatula
       & 
       A toy with a broad flat edge, used to splash people at the pool. \\
      \midrule
       stew skillet
      & 
      It is a skillet that you make stew on to make it taste really good. \\
       \midrule
       duck screen
       & A piece of artificial camouflage used by duck hunters.\\
      \bottomrule
    \end{tabular}
\end{center}

\begin{table*}[!ht]
    \centering
    \begin{tabular}{cccccc}
    \toprule
       \textbf{Compound type}  & \textbf{Example} & \textbf{Event} & \textbf{Perceptual} & \textbf{Environmental} & \textbf{N} \\
       \midrule
         art mod, art head & e.g. stew skillet & 93\% & 7\% & 0\% & N = 195\\
         nk mod, art head & e.g. stream wheel & 88\% & 12\%& 0\%& N = 157\\
         art mod, nk head & e.g. stew chickpea & 66\% & 34\% & 0\%& N = 157\\
         nk mod, nk head & e.g. stream vegetable & 15\%& 34\%& 51\%& N = 148\\
         \bottomrule
    \end{tabular}
    \caption{Distribution of metarelations per type of compounds from \citet{levin2019systematicity}.}
    \label{tab:dists}
\end{table*}

Overall, more exclusions happen in higher temperatures, and few-shot prompting drastically decreases the exclusion rate among generations. However, we also observe that few-shot-prompted generations have lower diversity in generated content, even with high temperatures. 

\subsection{Exclusion Patterns}

Examples of excluded model generations are presented in Table~\ref{tab:exclude}. Additional analyses of the excluded model generations reveal that GPT-3 makes both human-like and more machine-like mistakes. While some excluded generations are similar to excluded human definitions, such as failing to define the head of the compound or being vague, some exclusion cases are distinctly non-human. For example, the model sometimes defines the head and the modifier separately instead of as one word. 

Interestingly, the model seems to recognize the novelty of these compounds. The model often either prefaces the definition with equivalents of ``the term \texttt{X} is not a common one'' or states that ``\texttt{X} is not a real compound''.

One compound that the model often produces excluded generations for is \word{duck potato}. Specifically, the model attempt to implicitly define \word{duck} with water-related attributes, stating that duck potatoes are plants that grow in wet areas. Although the model sometimes explains its reasoning (e.g., ``the first thing that comes to mind when I see the word `duck' is the waterfowl. Therefore, I would think that a duck potato would be a potato that is found in the water, perhaps growing on the stem of a water plant.''), we exclude these explanations because they do not define a clear enough modifier--head relationship.

\subsection{Generation Stability}

We are interested in quantifying how different prompt types affect generation diversity. Therefore, we compute the Self-BLEU scores \cite{zhu2018texygen} of different generations per prompt type. For prompt type~1, the self-BLEU score is 58.24; for prompt type~2, the score is 52.49; for prompt type~3, however, the score is 77.39. 
Given that prompt types 2 and 3 differ only in the additional few-shot examples in 3, this suggests that few-shot prompting provides more information, hence higher stability, for this task.

\section{Experiment 2: Additional Details}\label{app:additional}

Although the novel compounds from Experiment~1 are unlikely to be present in GPT-3's training data, we generated additional novel compounds to try to factor out any possibility that GPT-3's generations are shaped by it having seen \citet{levin2019systematicity}'s materials during training.
The additional compounds do not have any lexical overlap with \citet{levin2019systematicity}. 

Furthermore, we observed in the previous experiment that the generations for \citet{levin2019systematicity}'s novel compounds are categorized primarily as the ``purpose'' metarelation subtype. Therefore, we also include compounds that, according to our intuitions, should have more varied metarelation subtypes. 

Here is a subset of our additional compounds with their intended metarelation subtypes:
\begin{center}
\setlength{\tabcolsep}{2pt}
    \begin{tabular}{cc}
        \toprule
           \textbf{Compound}  & \textbf{Intended Subtype} \\
             \midrule
           Bundt chair / tiger  & Borrowed\\
           Biden burrito / flower & Named after \\
           Wonder grass / popsicle & Value \\
           Elephant necktie / eggplant & Used-by \\
           Leaf necklace / stick & Distinctive part\\
           Monster bag / squid & Dimension \\
           \bottomrule
    \end{tabular}
    \label{tab:breadth_compounds}
\end{center}

We use the same temperature and prompt combinations as Experiment~1.

\subsection{Exclusion Rate}
Given that no large collection of human-created explanations are available for our additional novel compounds, we primarily use exclusion rate as the metric. Here, we modify the exclusion criteria from the previous experiment; we no longer exclude examples classified as ``other property'' by the majority of the raters, to explore the model's behavior in this category of noun compounds. The corresponding exclusion rates for each temperature--prompt combination are presented in Table~\ref{tab:exp2-exclude} in the main text.

\subsection{Generation Stability}
We observe some variation in generations overall, as there is more diversity in compound metarelations. For instance, for the compound \word{Biden burrito}, the following explanations are produced: 

\begin{enumerate}\setlength{\itemsep}{0pt}
    \item The Biden burrito would most likely refer to a burrito that is made in the style of Vice President Joe Biden. This could mean that the burrito is filled with Biden's favorite foods, or that it is made to look like Biden himself.
    \item The most likely interpretation of ``Biden burrito'' would be a burrito named after United States Vice President Joe Biden.
    \item Burrito that is eaten by Biden.
    \item The Biden burrito is a political term used to describe a 2020 United States presidential election campaign event in which then-Presidential candidate Joe Biden ate a burrito on a live stream.
\end{enumerate}

We can see that the generated explanations span across multiple metarelation subtypes and are relatively diverse for this specific instance due to its inherent ambiguity. 

Again, we compute the Self-BLEU scores of generations from each prompt type. Similar to what we observed in exclusion rate, the difference seems to diminish between different prompt types. The scores for Natural, Structured, and Few-shot prompts are 64.28, 54.06, and 65.68 respectively.

\section{Experiment 3: Additional Details}\label{app:experiment2}

\subsection{Proximity}\label{app:proximity}

When using WordNet for Mc, we identify nouns under a sufficiently specific cluster that is in the WordNet tree path to ensure path similarity, and locate the most common noun using unigram frequency.%
\footnote{\url{https://www.kaggle.com/datasets/rtatman/english-word-frequency}} We perform a similar operation for finding Mf instances, except using higher-level clusters to obtain low path-similarity nouns. For GloVe, we identify the most similar words and identify the first ones that differ in noun category as Dc instances, and use the least similar among the top 15,000 most similar words for Df instances. We use two different tools for this process because path similarity is only more informative when the two words have higher overlap in WordNet tree paths, and GloVe embeddings primarily capture semantic proximity without prioritizing noun categories.

\subsection{Exclusion Rates}

Similar to prior experiments, we prompt GPT-3 under 4 different temperatures: 0.2, 0.4, 0.7, and 0.9. We report the results from temperature 0.2 in the body of the paper because it is the best-performing. Here we include the exclusion rates of different head-modifier relation combinations from the other temperatures.

\begin{center}
    \centering
    \begin{tabular}{c *{4}{r} }
    \toprule
        \textit{Temperature: 0.4}\\
        \small{\underline{\textit{Head}}} & \textbf{Mc} & \textbf{Mf} & \textbf{Dc} & \textbf{Df}\\
        \midrule
        \small{\underline{\textit{Modifier}}} & & & &  \\
        \textbf{Mc} & 25.0 & 50.0 & 25.0 & 50.0 \\
        \textbf{Mf} &0.0  & 25.0 & 25.0 & 0.0  \\
        \textbf{Dc} & 50.0 & 75.0 & 25.0 & 33.3 \\
        \textbf{Df} & 100.0& 25.0 & 75.0 & 50.0 \\
        \bottomrule
    \end{tabular}
\end{center}

\begin{center}
    \begin{tabular}{c *{4}{r}}
    \toprule
    \textit{Temperature: 0.7}\\
        \small{\underline{\textit{Head}}} & \textbf{Mc} & \textbf{Mf} & \textbf{Dc} & \textbf{Df}  \\
        \midrule
        \small{\underline{\textit{Modifier}}} & & & &  \\
        \textbf{Mc} & 25.0 & 50.0 & 50.0 & 75.0 \\
        \textbf{Mf} &0.0  & 50.0 & 25.0 & 50.0 \\
        \textbf{Dc} & 25.0 & 50.0 & 50.0 & 50.0\\
        \textbf{Df} & 50.0 & 25.0 & 100.0 & 50.0 \\
        \bottomrule
    \end{tabular}
\end{center}

\begin{center}
    \centering
    \begin{tabular}{c *{4}{r}}
    \toprule
    \textit{Temperature: 0.9}\\
        \small{\underline{\textit{Head}}} & \textbf{Mc} & \textbf{Mf} & \textbf{Dc} & \textbf{Df}  \\
        \midrule
        \small{\underline{\textit{Modifier}}} & & & &  \\
        \textbf{Mc} & 25.0 & 50.0 & 50.0 & 75.0 \\
        \textbf{Mf} &0.0  & 25.0 & 25.0 & 25.0 \\
        \textbf{Dc} & 50.0 & 50.0 & 50.0 & 25.0\\
        \textbf{Df} & 75.0 & 25.0 & 25.0 & 50.0 \\
        \bottomrule
    \end{tabular}
\end{center}

\subsection{Full Materials}\label{app:materials}
\subsubsection{Basis Compound Definitions}

\begin{table*}[!ht]
\centering
\begin{tabular}{l l l l l} 
\toprule
\textbf{Baseline Word} & \textbf{Mc Word} & \textbf{Mf Word} & \textbf{Dc Word} & \textbf{Df Word}  \\
\midrule
kitchen & bedroom & trunk & vegetable & fur  \\
knife & scissors & magazine & metal & shrubbery  \\
tree & bush & tomato & garden & blender  \\
frog & reptile & doe & slipper & satin  \\
strawberry & banana & bamboo & shortcake & overcoat  \\
cookie & pancake & table & egg & oak  \\
coffee & cider & medicine & fruit & kangaroo  \\
bean & nut & flower & chili & gear \\
\bottomrule
\end{tabular}
\caption{Each baseline component of the compounds and their corresponding generated words based on matching or different noun category (M/D) and close or far semantic similarity (c/f)}
\label{tab:full_alien}
\end{table*}

We report the list of definitions we use as a part of the basis compound definitions.

\begin{center}
\setlength{\tabcolsep}{2pt}
\begin{tabular}{P{2cm} P{5cm}}
\toprule
    \textbf{Compound} & \textbf{Definition} \\
    \midrule
    kitchen knife & A ``kitchen'' is a room equipped for preparing meals. A ``knife'' is edge tool used as a cutting instrument; has a pointed blade with a sharp edge and a handle. A ``kitchen knife'' is a knife used in cooking . \\
    tree frog & A ``tree'' is a tall perennial woody plant having a main trunk and branches forming a distinct elevated crown. A ``frog'' is any of various tailless stout-bodied amphibians with long hind limbs for leaping; semiaquatic and terrestrial species. A ``tree frog'' is a frog that lives in trees.\\
    strawberry cookie & A ``strawberry'' is sweet fleshy red fruit. A ``cookie'' is any of various small flat sweet cakes. A ``strawberry cookie'' is a cookie made with strawberries. \\
    coffee bean & A ``coffee'' is a beverage consisting of an infusion of ground coffee beans. A ``bean'' is any of various edible seeds of plants of the family Leguminosae used for food. A ``coffee bean'' is a bean used to make coffee.\\
    \bottomrule
\end{tabular}
\end{center}

\subsubsection{Definitions of Derived Compounds}
The full list of derived compounds from the four basis compounds are in Table~\ref{tab:full_alien}. To reiterate, these new compounds created only inform us of what definitions should be a part of the prompt, and the specific words are not shown to the model in the prompt.

For these derived compounds, we first identify the corresponding WordNet definition, then truncate and modify these definitions to prevent too much leakage of information so that the model should not be able to pinpoint the exact word from the prompt. The complete list of compound components used in the experiments, including both the original and the modified definitions, is presented below.\\

\noindent 
\begin{tabular}{@{} p{\linewidth} @{}}
    \toprule
     \textbf{Basis word: kitchen}\\
     \bottomrule
\end{tabular}\\

\textbf{bedroom: }
\begin{itemize}
    \item Original: a room used primarily for sleeping
    \item Modified: a room with furniture
\end{itemize}

\textbf{trunk:}
\begin{itemize}
    \item Original: compartment in an automobile that carries luggage or shopping or tools
    \item Modified: compartment in an automobile
\end{itemize}

\textbf{vegetable:}
\begin{itemize}
    \item Original: edible seeds or roots or stems or leaves or bulbs or tubers or nonsweet fruits of any of numerous herbaceous plant
    \item Modified: edible plant
\end{itemize}

\textbf{fur:}
\begin{itemize}
    \item Original: dense coat of fine silky hairs on mammals
    \item Modified: part of a mammals body
\end{itemize}

\noindent \begin{tabular}{@{} p{\linewidth} @{}}
    \toprule
     \textbf{Basis word: knife}\\
     \bottomrule
\end{tabular}\\

\textbf{scissors: }
\begin{itemize}
    \item Original: an edge tool having two crossed pivoting blades
    \item Modified: a tool with blades
\end{itemize}

\textbf{magazine:}
\begin{itemize}
    \item Original: a periodic publication containing pictures and stories and articles of interest to those who purchase it or subscribe to it
    \item Modified: a publication with pictures
\end{itemize}

\textbf{metal:}
\begin{itemize}
    \item Original: any of several chemical elements that are usually shiny solids that conduct heat or electricity and can be formed into sheets etc.
    \item Modified: any shiny solid
\end{itemize}

\textbf{shrubbery:}
\begin{itemize}
    \item Original: a collection of shrubs growing together
    \item Modified: a collection of plants
\end{itemize}

\noindent \begin{tabular}{@{} p{\linewidth} @{}}
    \toprule
     \textbf{Basis word: tree}\\
     \bottomrule
\end{tabular}\\

\textbf{bush:}
\begin{itemize}
    \item Original: a low woody perennial plant usually having several major stems
    \item Modified: a low woody plant
\end{itemize}

\textbf{tomato:}
\begin{itemize}
    \item Original: mildly acid red or yellow pulpy fruit eaten as a vegetable
    \item Modified: pulpy fruit
\end{itemize}

\textbf{garden:}
\begin{itemize}
    \item Original: a yard or lawn adjoining a house
    \item Modified: an outside area
\end{itemize}

\textbf{blender:}
\begin{itemize}
    \item Original: an electrically powered mixer with whirling blades that mix or chop or liquefy foods
    \item Modified: an appliance
\end{itemize}

\noindent \begin{tabular}{@{} p{\linewidth} @{}}
    \toprule
     \textbf{Basis word: frog}\\
     \bottomrule
\end{tabular}\\

\textbf{reptile:}
\begin{itemize}
    \item Original: any cold-blooded vertebrate of the class Reptilia including tortoises, turtles, snakes, lizards, alligators, crocodiles, and extinct forms
    \item Modified: cold-blooded vertebrate
\end{itemize}

\textbf{doe:}
\begin{itemize}
    \item Original: mature female of mammals of which the male is called `buck'
    \item Modified: mature mammals in the forest
\end{itemize}

\textbf{slipper:}
\begin{itemize}
    \item Original: low footwear that can be slipped on and off easily
    \item Modified: low footwear
\end{itemize}

\textbf{satin:}
\begin{itemize}
    \item Original: a smooth fabric of silk or rayon
    \item Modified: a fabric
\end{itemize}

\noindent \begin{tabular}{@{} p{\linewidth} @{}}
    \toprule
     \textbf{Basis word: strawberry}\\
     \bottomrule
\end{tabular}\\

\textbf{banana:}
\begin{itemize}
    \item Original: any of several tropical and subtropical treelike herbs of the genus Musa having a terminal crown of large entire leaves and usually bearing hanging clusters of elongated fruits
    \item Modified: a tropical yellow fruit 
\end{itemize}

\textbf{bamboo:}
\begin{itemize}
    \item Original: woody tropical grass having hollow woody stems; mature canes used for construction and furniture
    \item Modified: woody tropical plant
\end{itemize}

\textbf{shortcake:}
\begin{itemize}
    \item Original: very short biscuit dough baked as individual biscuits or a round loaf
    \item Modified: short biscuit
\end{itemize}

\textbf{overcoat:}
\begin{itemize}
    \item Original: a heavy coat worn over clothes in winter
    \item Modified: a piece of clothing
\end{itemize}

\noindent \begin{tabular}{@{} p{\linewidth} @{}}
    \toprule
     \textbf{Basis word: cookie}\\
     \bottomrule
\end{tabular}\\

\textbf{pancake:}
\begin{itemize}
    \item Original: a flat cake of thin batter fried on both sides on a griddle
    \item Modified: a flat cake 
\end{itemize}

\textbf{table:}
\begin{itemize}
    \item Original: a piece of furniture having a smooth flat top that is usually supported by one or more vertical legs
    \item Modified: a piece of furniture
\end{itemize}

\textbf{egg:}
\begin{itemize}
    \item Original: oval reproductive body of a fowl (especially a hen) used as food
    \item Modified: food from a farm animal
\end{itemize}

\textbf{oak:}
\begin{itemize}
    \item Original: a deciduous tree of the genus Quercus
    \item Modified: a tree
\end{itemize}

\noindent \begin{tabular}{@{} p{\linewidth} @{}}
    \toprule
     \textbf{Basis word: coffee}\\
     \bottomrule
\end{tabular}\\

\textbf{cider:}
\begin{itemize}
    \item Original: a beverage made from juice pressed from apples
    \item Modified: a beverage made from juice
\end{itemize}

\textbf{medicine:}
\begin{itemize}
    \item Original: something that treats or prevents or alleviates the symptoms of disease
    \item Modified: something that treats sickness
\end{itemize}

\textbf{fruit:}
\begin{itemize}
    \item Original: the ripened reproductive body of a seed plant
    \item Modified: a part of a plant
\end{itemize}

\textbf{kangaroo:}
\begin{itemize}
    \item Original: any of several herbivorous leaping marsupials of Australia and New Guinea having large powerful hind legs and a long thick tail
    \item Modified: a mammal
\end{itemize}

\noindent \begin{tabular}{@{} p{\linewidth} @{}}
    \toprule
     \textbf{Basis word: bean}\\
     \bottomrule
\end{tabular}\\

\textbf{nut:}
\begin{itemize}
    \item Original: usually large hard-shelled seed
    \item Modified: hard-shelled seed
\end{itemize}

\textbf{flower:}
\begin{itemize}
    \item Original: a plant cultivated for its blooms or blossoms
    \item Modified: a colorful plant
\end{itemize}

\textbf{chili:}
\begin{itemize}
    \item Original: ground beef and chili peppers or chili powder often with tomatoes and kidney beans
    \item Modified: a southern soup
\end{itemize}

\textbf{gear:}
\begin{itemize}
    \item Original: a toothed wheel that engages another toothed mechanism in order to change the speed or direction of transmitted motion
    \item Modified: a wheel
\end{itemize}

\end{document}